\documentclass[letterpaper, 10 pt, conference]{ieeeconf}  % Comment this line out if you need a4paper

\IEEEoverridecommandlockouts                              % This command is only needed if
\usepackage{graphics} % for pdf, bitmapped graphics files
\usepackage{epsfig} % for postscript graphics files
\usepackage{mathptmx} % assumes new font selection scheme installed
\usepackage{amsmath} % assumes amsmath package installed
\usepackage{amssymb}  % assumes amsmath package installed
\usepackage{tabularx}
\usepackage{multirow}
\usepackage{color}
\usepackage{url}
\usepackage{breqn}
\newcommand{\junk}[1]{}
\usepackage{algorithm2e}
\usepackage{dblfloatfix} 
\usepackage[export]{adjustbox}
\usepackage{tabulary,booktabs}
\title{\LARGE \bf
Improved Reinforcement Learning Coordinated Control of a Mobile Manipulator using Joint Clamping
}

\author{Denis Hadjivelichkov$^{1}$, Kostas Vlachos$^{2}$, Dimitrios Kanoulas$^{1}$% <-this % stops a space
\thanks{$^{1}$
University College London, 
{\tt\small {denis.hadjivelichkov.19, d.kanoulas}@ucl.ac.uk}}
\thanks{$^{2}$
University of Ioannina,
{\tt\small kostaswl@cse.uoi.gr}}}%

\begin{document}

\maketitle
\thispagestyle{empty}
\pagestyle{empty}

\begin{abstract}
%Many robotic path planning problems are continuous, stochastic and high-dimensional. The ability of a mobile manipulator to coordinate its base and manipulator in order to control its whole-body is particularly challenging when self- and environment- collision avoidance is required based on partial observations. Modern Reinforcement Learning techniques have the potential to solve such problems through their ability to generalize. In this paper, we identify directions for further improvement of a state-of-the-art reinforcement learning system for mobile manipulator Whole-Body Control (WBC), and show a modification of its agent and handcrafted reward function that lead to higher success rates.
Many robotic path planning problems are continuous, stochastic, and high-dimensional. The ability of a mobile manipulator to coordinate its base and manipulator in order to control its whole-body online is particularly challenging when self and environment collision avoidance is required. %based on partial observations. 
%Modern 
Reinforcement Learning techniques have the potential to solve such problems through their ability to generalise over environments. 
%In this paper, 
We study joint penalties and joint limits of a state-of-the-art mobile manipulator whole-body controller that uses LIDAR sensing for obstacle collision avoidance.  We propose directions to improve the reinforcement learning method.
%regarding the shaped reward function characteristics
Our agent achieves significantly higher success rates than the baseline in a goal-reaching environment and it can solve environments that require coordinated whole-body control which the baseline fails. 
\end{abstract}

\section{INTRODUCTION}\label{Sec:intro}
Mobile robots have a plethora of applications ranging from warehouse services, through oil rig inspections, to emergency interventions~\cite{Rajana2018, Bengel2009, Rehak2013}. Modern robots require both high mobility and accurate manipulation to traverse collision-free paths while performing their tasks, which can be achieved via \emph{mobile manipulators}. By applying Whole-Body Control (WBC), the base and manipulator movements of mobile robots coordinate to improve the efficiency of the system. 

Classical WBC methods include the use of kinematic, velocity, and impedance controllers, model predictive controllers, and combinations thereof in advanced adaptive control strategies~\cite{Moro2018, Logothetis2018, Kim2019}. They have been shown to work well in many environments while also providing stability guarantees.  Reinforcement Learning (RL) methods have shown great promise in their ability to compete with and potentially overcome classical methods in many robotic problems as they can work with complex inputs~\cite{jaderberg2016} and learn complex task solutions~\cite{levine2018reinforcement}. Once trained, RL agents can execute policies online, bringing down total mission times.

The recent state-of-the-art works on RL for mobile manipulator WBC have focused on goal reaching scenarios.  While they show that their solutions are quicker than traditional methods they still underperform them in terms of success rate and impose big limitations on the robots and their environments, such as limited DoF~\cite{kindle2020wholebody} and simplistic tasks~\cite{Wang2020}. 

\begin{figure}[!t]
    \centering
    \includegraphics[width=0.5\columnwidth]{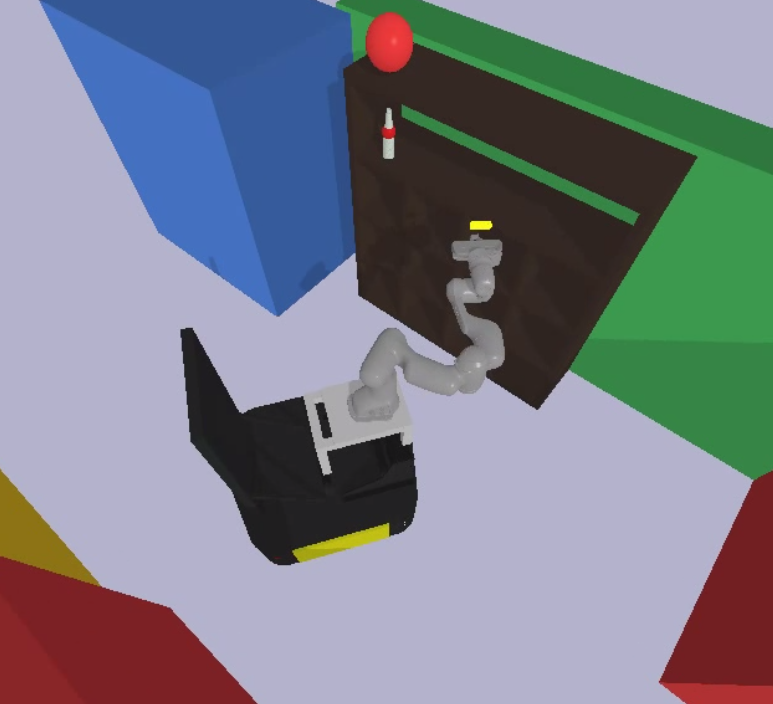}
    \caption{The simulated mobile manipulator, composed by an omnidirectional mobile base and a 7DoF arm.}
    \label{fig:intro:simulated-robot}
\end{figure}

In this paper, we examine the trained WBC behaviour of a state-of-the-art baseline agent trained with a shaped reward.  The robot agent is comprised of an omnidirectional base and a 7 DoF manipulator simulated in PyBullet (see Fig.~\ref{fig:intro:simulated-robot}).  We determine  potential causes of sub-optimality in the baseline - frequent early episode termination due to reaching joint limits and consistently folded arm. We show that clamping the joint instead of a joint-limit penalty in the the reward improves the model's performance significantly and allows it to reach the goal much closer.

The summarised contributions of this paper are:
(i) Identifying issues and potential improvements of a state-of-the-art method;
(ii) Showing our method leads to higher success rates than the baseline and solves an environment requiring whole-body control, which the baseline fails;
(iii) Evince our method's ability to generalise in an unseen environment.

\section{RELATED WORK}\label{Sec:rw}
In this section, we present traditional and reinforcement learning approaches to whole-body control, justifying its use.

\textbf{Traditional Approaches:} Traditional approaches to implementing WBC include the use of kinematic and dynamic controllers~\cite{Khatib1987, bauza2017probabilistic, Wu2019}. Their advantage is that the current understanding of physical systems is refined and works well on fully actuated robots.  Most methods focus on WBC for quadrupeds~\cite{Gennaro2020, Bellicoso2017, Morlando2021}, humanoids~\cite{Dietrich2012, Mansard2009sot, hoffman2018icub, Rocchi2015opensot, Laurenzi2019}, animaloids~\cite{Rolley2018, Raghavan2019, Raghavan2020}, or mobile manipulators~\cite{Dietrich2012, Liu2020}. Model Predictive Control methods are popular with works such as Minniti et al.~\cite{Minniti2019} showing success in WBC pose-tracking and interaction tasks. Recent works focus on non-linear strategies, such as Hierarchical Quadratic Programming~\cite{Kim2019}, and non-linear model predictive control~\cite{Logothetis2018}. While some methods such as Operational Space Control can solve tasks with optimality and continuity in real-time, most traditional methods require large offline computation. Moreover, the methods for mobile manipulators are often based on simplified models of the robot which sometimes results in control solutions that are limiting the its agility.

\textbf{Reinforcement Learning Approaches:}
Reinforcement learning approaches offer a framework that is transferable to different tasks and robots, able to work online with scaling complexity, in a trade off with the limited prior information that it can use and long training that is often required for good performance. However, current methods still use application-specific architectures and rarely generalize to multi-task scenarios~\cite{Claude2012}.  RL methods have successfully taught robots dexterous vision-based manipulation tasks~\cite{Andrychowicz2020, openai2019solving, kalashnikov2018qtopt, julian2020efficient} and navigation tasks~\cite{Hester2011realtimeRL, francis2019longrange}.

Most research is also focused on legged robots~\cite{Li2018,Hwangbo2019,Yang2018, Lober2016}.
Wang et al.~\cite{Wang2020} integrate the state-of-the-art RL algorithms with visual perception for WBC and propose an efficient framework for decoupling of visual perception from control, which enables easier sim-to-real transfer.  However, the used environment is simple, consisting of a table in front of a robot. Kindle et al.~\cite{kindle2020wholebody} use a Proximal Policy Optimization (PPO) based agent to train end-to-end whole-body control policies for obstacle avoidance and tested on a real mobile manipulator achieving state-of-the-art results. Their model makes use of Automatic Domain Randomization and Continuous Learning to guide the agent toward a solution in a custom reach-and-grasp environment.  A hand-crafted reward function is defined with components for collision, joint limits, safety distance, optimal path following and time. These recent works show sub-optimal performance, worse than comparable traditional methods.
\section{BACKGROUND}\label{Sec:background}
We consider a standard RL framework, which includes an agent interacting with an environment via actions and observations.  Environment rewards are fed into an RL learning algorithm, 
which optimises the agent's policy and thus creates a feedback loop.  The problem focuses on goal-reaching environments in which a success is defined as the uninterrupted holding of the robot agent's end-effector within a given tolerance distance from the goal. The environments' state space consists of front and rear LIDAR scans, arm joint positions, arm joint and base velocities, and the goal location in the end-effector frame, while action space consists of joint and base accelerations. Both LIDAR observations and joint actions are limited to 2D planes.   In this section, we discuss he state-of-the-art baseline~\cite{kindle2020wholebody} used for our experiments.

\subsubsection{Reward}
The baseline's reward function is handcrafted, encouraging the agent to learn to imitate a traditional path planning method and complete the task quicker, while discouraging it for moving close to objects. The reward has three termination cases: collision, timeout, and reaching joint limits. Finally, it also introduces an accumulation term that prevents the agent's exploitation of the reward.

\subsubsection{Agent}
\begin{figure}[!t]
    \centering
    \includegraphics[width=0.85\columnwidth]{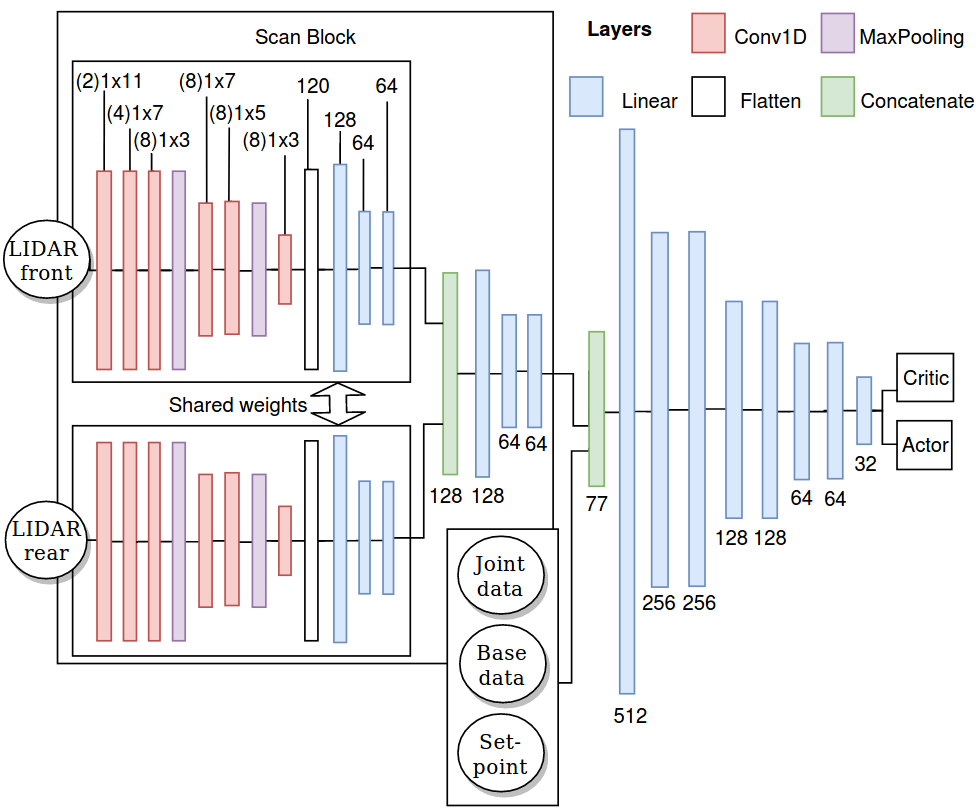}
    \caption{Agent's network architecture.}
    \label{fig:bg:agent-architecture}
\end{figure}

The architecture of the agent is based on PPO with modified layers as depicted in Fig.~\ref{fig:bg:agent-architecture}. The two LIDAR scans are compressed via a separate scan block before being processed with the rest of the inputs in a network of fully connected layers. The agent produces a discretized policy for each action and its respective value.
\section{METHOD}\label{Sec:method}
\begin{figure}[!h]
    \centering
    \begin{minipage}{0.405\columnwidth}
        \centering        \includegraphics[width=0.99\columnwidth]{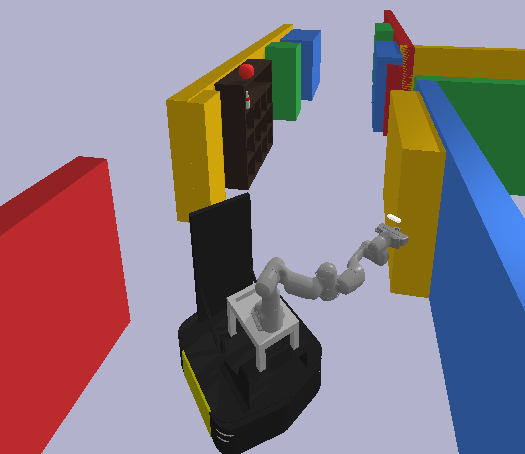}
    \end{minipage}%
    \begin{minipage}{0.3\columnwidth}
        \centering
        \includegraphics[width=0.99\columnwidth]{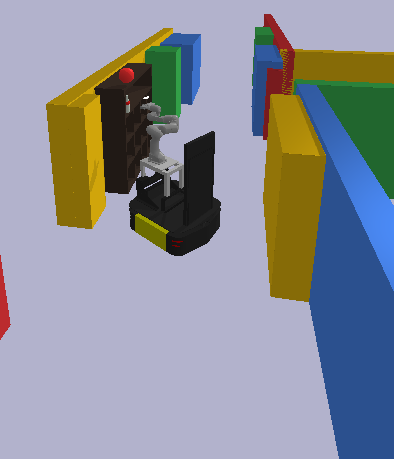}
    \end{minipage}
    \caption{The robot controlled with the baseline agent at start (left) and end (right) of reaching task. Note the folded arm in the second image.}
    \label{fig:reaching-jl}
\end{figure}

\begin{figure*}[h]
    \centering
    \begin{minipage}{.43\columnwidth}
        \centering
        \includegraphics[width=0.99\columnwidth]{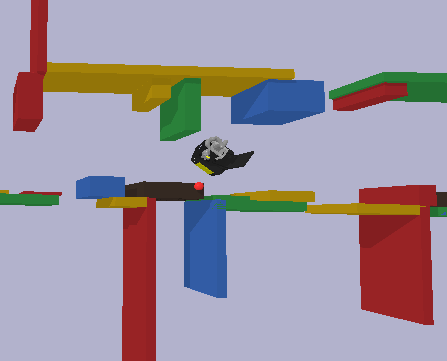}
    \end{minipage}
    \begin{minipage}{.59\columnwidth}
        \centering
        \includegraphics[width=0.99\columnwidth]{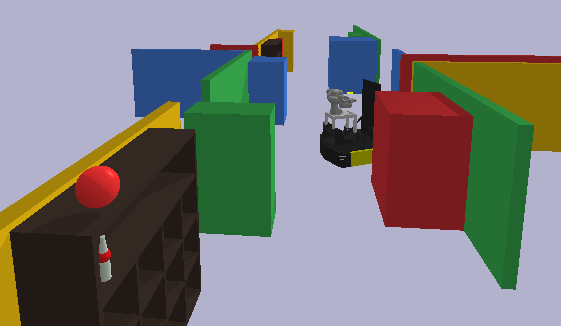}
    \end{minipage}
    \begin{minipage}{0.345\columnwidth}
        \centering
       \includegraphics[width=0.99\columnwidth]{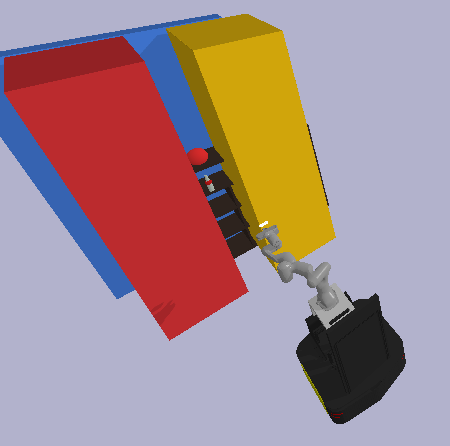}
    \end{minipage}
    \begin{minipage}{0.342\columnwidth}
        \centering
      \includegraphics[width=0.99\columnwidth]{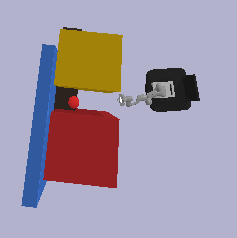}
    \end{minipage}
    \caption{The \emph{Corridor-env} (left two) requires optimal path planning to be solved, but allows for the base and manipulator to move independently. The local \emph{Gap-env} (right two) cannot be solved without coordinated control with the robot inserting its manipulator in the tunnel to approach the goal.}
    \label{fig:method:custom-environment}
\end{figure*}

To understand the low success rate of the baseline in comparison with traditional methods, we analysed the behaviour and performance of the agent after training. It was observed that the majority of episodes terminate due to the robot arm reaching its joint limits. We further noticed that this is a behaviour that can be limited explicitly instead of penalising and terminating the reinforcement learning agent.

While it was expected that the optimal solution would be for the robot to move toward the goal with a folded arm and unfold it while it is reaching the goal position, it was observed that the robot folds the arm in the beginning and does not change it throughout the whole run, as shown in Fig.~\ref{fig:reaching-jl} as well as the real robot experiments~\cite{kindle2020youtube}. The cause of this could be partially explained by the custom environment itself, which does not explicitly require WBC in order to be solved. Additionally, the used reward function itself places more weight on optimal path following penalties than on timing penalties - following the optimal end-effector path is simpler when the manipulator is folded, because the end-effector is close to the base point of rotation.  We attempt to address some of these drawbacks in this section.

\textbf{Improved Environments:} Two environments are used in our validations: a narrow corridor environment for comparison with state-of-the art and a new environment that cannot be solved without WBC.

The first environment,  adapted from~\cite{kindle2020wholebody}, consists of a narrow corridor of variable length, containing random avoidable obstacles and a randomly placed goal location (See Fig.~\ref{fig:method:custom-environment}).  It requires the agent to plan its path and navigate around the obstacles toward the goal. We refer to this environment as \emph{Corridor-env}.

However, its goal can be directly reached by folding the arm and performing only mobile base collision-free navigation, thus does not require WBC.

We introduce a new environment (referred as \emph{Gap-env}) which consists of narrow passages with the goal end-effector pose being reachable only with coordination between the base and manipulator
(See Fig.~\ref{fig:method:custom-environment}).  The width of the gap is very narrow and a small deviation of the arm or base during insertion would cause a collision, thus the task is difficult to solve without coordinated control.  Two variants are used: In \emph{Gap-env-train}, random uniform noise is added to initial joint angles, and orientation and position of the goal relative to the robot spawn location; In \emph{Gap-env-test}, the tunnel gap width and length, as well as the goal placement relative to the tunnel is also randomly initialized to ensure that the testing scenarios are unseen by the agent. 

In all the environments, a success is defined as the uninterrupted holding of the robot agent's end-effector within a given tolerance distance from the goal. Automatic Domain Randomization (ADR) is used to gradually adapt the complexity of the environment and guide the agent toward a solution. This is done by increasing or decreasing the acceptable tolerance distance to the goal depending on the agent's recent success rate. Via ADR, the tolerance distance to the goal is dynamically changed. The state space consists of 2D front and rear LIDAR scans, arm joint positions, arm joint and base velocities, and the goal location in end-effector frame. The action space is comprised of mobile base and arm joint accelerations. Given the complexity of the problem, its dimensionality is reduced to planar movements of the arm.

\textbf{Joint Clamping Method:}\label{sec:method:joint-limit-extension}
When training the baseline agent in the corridor environment, it was noticed that most of the episodes end due to joint limit termination.  However, joint limits can easily be enforced by setting a manual limit (clamping) to the joint positions based on the robot's hardware limits with appropriate tolerance to protect the robot. Likewise, when training the baseline in the gap environment, it doesnt approach the goal, rather stays near the gap and oscillates. This is believed to be due to the baseline's safety margin penalty, which encourages the robot to keep its distance from all objects. We believe that a collision termination penalty is sufficient in teaching that behaviour. Thus, our modified reward function is as follows:

\begin{equation}\label{eq:handcrafted-reward-function-reduced}
\begin{aligned}
    r_t &= 
    w_t . \cfrac{\tau}{T_t}  
    +
    w_{pd} \Delta d_{pd} + w_{pt} \cfrac{\Delta d_{pt}}
    { d_{pt,init}}+
    %w_{sm} |v|\tau 
    %(1-\min(1, d_{sm}/d_{th}) )
    %\\&+
    w_{ht}\cfrac{\tau}{T_h} 
    \\&+ w_{hd} (1-\min(1,d_g/d_h))\cfrac{\tau}{T_h} 
    -
    I_h
    +
    D_c + D_h
\end{aligned}
\end{equation}
where $w_t$ is the time penalty parameter, $\tau$ is the step time, and $T_t$ is the total time before episode timeout. This timeout reward encourages quicker task completion.  An optimal path towards the goal is computed via Harmonic Potential Field (HPT). The goal distance reward penalises for deviation from the HPT $\Delta d_{pd}$ by $w_{pd}$ and rewards movement along the path $\Delta d_{pt}$ normalized by the total path $d_{pt,init}$ by $w_{pt}$. Furthermore, $w_{ht}$ is the reward for each time-step that the end-effector is within tolerance distance $d_h$ of the goal point and $w_{hd}$ is the reward for minimizing the distance to the goal position applied only when the distance to the goal $d_g$ is smaller than the tolerance distance $d_h$.  Both of these rewards are normalized for the holding time threshold $T_h$ after which the task is done. $I_h$ is the accumulated holding reward which is subtracted if the end-effector leaves the tolerance sphere in order to prevent exploitation of the reward. Finally, $D_c$ is a collision penalty, and $D_h$ is the reward for sustained holding time $T_h$.  The last two rewards end the current episode.

This reward function allows the agent to actuate the robot safely without hindering its learning and addresses the two issues encountered when running the baseline. The joint clamping is enforced programmatically, based on the robot's joint limits. The baseline agent is trained in with these modifications and compared with the standard baseline for several values of goal tolerance distance in Sec.~\ref{Sec:Exp-Baseline}. 

\textbf{Validation Setup}\label{sec:method:baseline-setup}
We use a mobile manipulator robot comprised of a omnidirectional base and a 7 DoF arm manipulator.  All simulations are done using PyBullet 2.8~\cite{coumans2020}, while a high performance computing cluster is used for training. Agent parameters are shown in the appendix.

Our method and its compared baseline are trained in \emph{Corridor-env} on $32$ parallel workers for a total of $60$M training steps. The final success rate, counted as number of successes over $100$ episodes, is compared in Sec.~\ref{Sec:Exp-Baseline}.   We train the agents in \emph{Gap-env-train}
for $30$M steps with $16$ parallel workers. The agent is then evaluated in \emph{Gap-env-test} and the results are reported in Sec.~\ref{Sec:Exp-WBC}. In both environments, the ADR is gradually adapting the tolerance distance in the range $[0.5;0.05]$. 
\section{RESULTS}\label{Sec:exp}
We explore the following questions: (i) How does our method compare to the baseline? (ii) Is the new agent able to perform well on a task requiring Whole-Body Control? (iii) Is the agent generalizable to new environments?

\subsection{Validation on Corridor Environment}\label{Sec:Exp-Baseline}

\begin{figure}[!th]
    \centering
    \begin{minipage}{0.34\columnwidth}
        \centering
        \includegraphics[width=0.99\columnwidth]{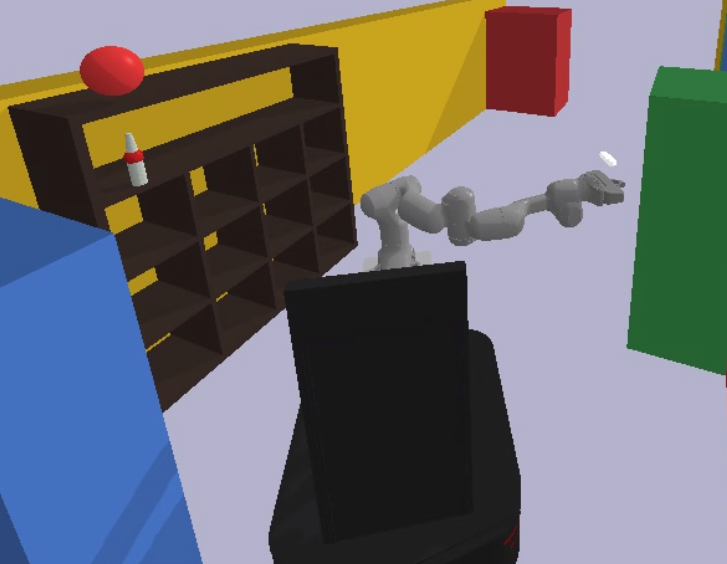}
    \end{minipage}%
    \begin{minipage}{0.34\columnwidth}
        \centering
        \includegraphics[width=0.99\columnwidth]{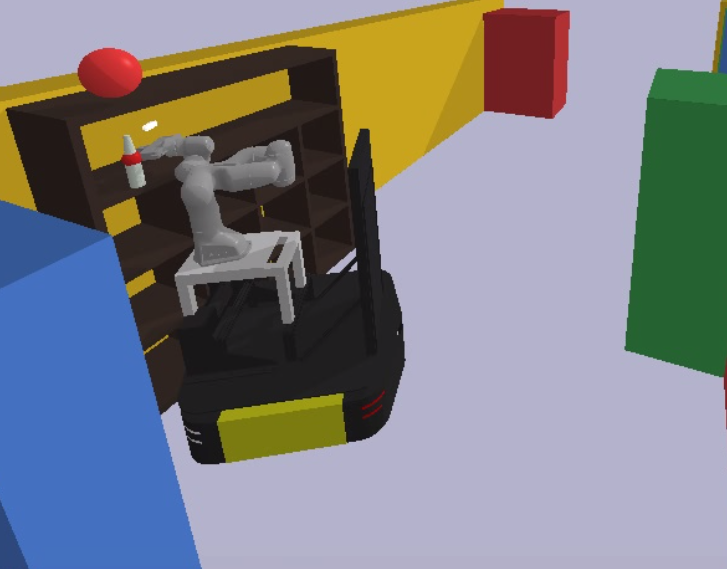}
    \end{minipage}
    \caption{Our agent at start (left) and end (right) of reaching task.}
    \label{fig:reaching-nojl}
\end{figure}

\begin{table}[!t]
    \centering 
    \begin{tabular}{|c||c|c|c|c|}
        \hline
        Tolerance Dist (m) & 0.5 & 0.2 & 0.1 & 0.07 \\\hline\hline
        Baseline &  
            72\% & 65\%& 40\% & 0\% \\\hline
        Joint-Clamping Method (ours) & 
            \textbf{73}\% & \textbf{73}\%& \textbf{63}\% & \textbf{24}\%\\\hline
        %Baseline (reported) \cite{kindle2020wholebody} & 
        %    - & - & - & \textbf{55}\% \\\hline
    \end{tabular}
    \caption{Success rates in \emph{Corridor-env} against tolerance distances}
    \label{tab:baselinerewsuccrate}
\end{table}

\begin{table}[!t]
    \centering 
    \begin{tabular}{|c||c|c|}
        \hline
        Environment & Gap-env-train & Gap-env-test \\\hline\hline
        Baseline & fail & fail \\\hline
        Joint Clamping Method (ours)  & 81\% & 76\% \\\hline 
    \end{tabular}
    \caption{Success rates in \emph{Gap-env} with $0.05m$ tolerance distance.}
    \label{tab:localenv-succrate}
\end{table}

\begin{figure}[!t]
\centering
\begin{minipage}{0.85\columnwidth}
    \centering
    \includegraphics[width=0.99\columnwidth]{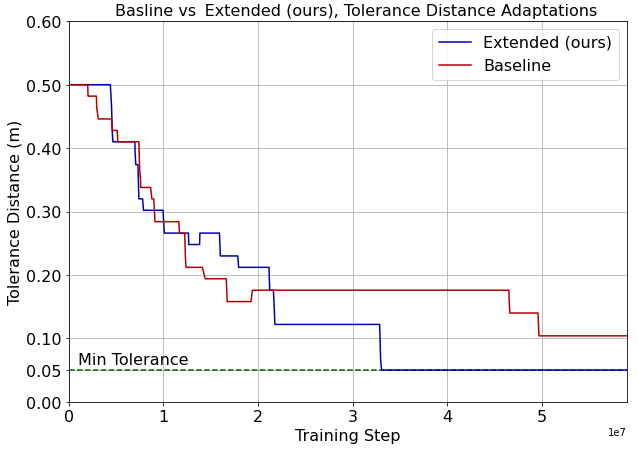}
\end{minipage}
\begin{minipage}{0.85\columnwidth}
    \centering
    \includegraphics[width=0.99\columnwidth]{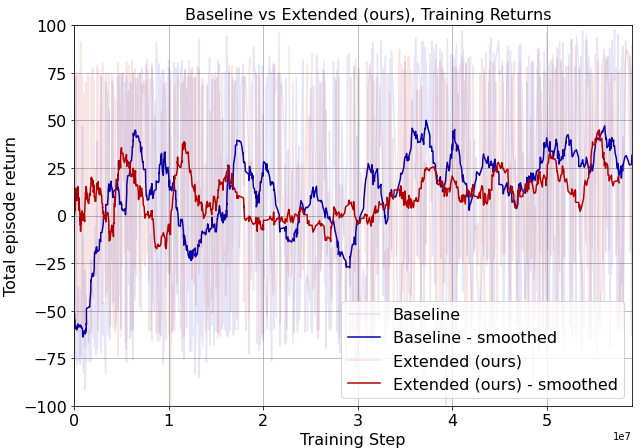}
\end{minipage}
    \caption{\textbf{(left)} Adaptation of tolerance distance per step for both reward settings. 
    %Blue: baseline without joint penalty (ours); Red: baseline with joint penalty. 
    \textbf{(right)} Total returns per episode plotted against the episode termination step.  Running mean smoothing of $0.95$ is used. Note that due to the use of ADR, the training rewards are fairly similar}
    \label{fig:adrboth}
\end{figure}

The original baseline agent with a joint limit penalty and our modified agent with clamped joint limits (Eq.~\ref{eq:handcrafted-reward-function-reduced}) are ran with $32$ workers for $60$M steps. This process took $48$ hours to finish. The trained models were tested in \emph{Corridor-env} with fixed goal tolerance distances in the range of $0.5$ to $0.07$. 

Running the baseline, we managed to achieve $72\%$ for the highest tolerance distance ($0.5$m), while the success rate was significantly decreasing with the tolerance distance dropping.  The minimum successful tolerance distance was $0.1$m. The resulting success rates of our agent, shown in the lower row of Table~\ref{tab:baselinerewsuccrate}, are noticeably higher than the baseline performance with standard reward, especially when the tolerance distance is decreasing.

This difference in performance can be further explained by the difference of ADR tolerances shown in Fig.~\ref{fig:adrboth}. For the baseline, the ADR tolerance distance reached at $60$M steps is $0.1$, not reaching the lowest distance of $0.05$. In comparison, our agent successfully adapted to the lowest tolerance distance $20$M steps before the training ended. This indicates that with the modified reward, the agent learns quicker and better. The training returns of the original and modified baselines are shown in Fig.~\ref{fig:adrboth}.  For the modified agent in the environment with tolerance distance fixed to $0.07m$, it is found that $52\%$ of the  unsuccessful episodes terminate due to collision, while $48\%$ terminate due to timeout. The distance from the end-effector to the goal at the end of unsuccessful episodes is on average $0.11m$.  While the joint limit modification shows an increase in success rate, the folded manipulator behaviour is still observed (see Fig.~\ref{fig:reaching-nojl}).

\subsection{Whole-Body Control Task}\label{Sec:Exp-WBC}
Training the agent locally in \emph{Gap-env}, which includes narrow tunnels where only the arm can fit, forces the simultaneous coordination between the arm and the mobile base as a WBC.  With a $0.05$m tolerance distance to the goal, the success rate of the training is $81\%$, shown in Table~\ref{tab:localenv-succrate}.  As can be observed from the success rates in the table, our agent successfully generalises to unseen variants of the training environment, with a drop of only 5\% in success rate.  In a typical episode, the robot is observed moving towards the goal, while adjusting its manipulator for tunnel-entry, as expected from a WBC solution. In failed episodes, it is seen that the robot often reaches the goal within less than $0.05m$, however it backs off and re-approaches several times until the episode terminates due to timeout.  Note that the original baseline method was not able to solve such environments based on its handcrafted reward function.  

%\section{DISCUSSION OF OTHER ATTEMPTS}\label{Sec:discussion}
%While trying to improve the baseline, we had several unsuccessful attempts, two of which are briefly discussed in this section: SAC backbone and LIDAR auto-encoding. In preliminary testing on simple robotic tasks, we identified that Soft Actor-Critic (SAC) RL methods~\cite{haarnoja2018sac} have comparable performance to PPO, while training quicker. For this reason, we implemented a modified baseline that uses SAC instead of PPO to form its policy.  However, the SAC agent was able to reach significantly less steps than the baseline and consistently failed the environments during evaluation.

%Another attempt was to reduce the computational time of the system by replacing the LIDAR pre-processing layers with a pre-trained LIDAR auto-encoder. Although the auto-encoding error was small, the respective success rate achieved was only $10\%$ at tolerance distance of $0.5m$. This shows that while the auto-encoder was able to preserve most of the information, the convolutional layers that it replaced in the agent were learning more useful features.
\section{CONCLUSIONS AND FUTURE WORK}\label{Sec:concl}
This work presents an RL method for Whole-body Control of a mobile manipulator that improves on the state of the art.  Our shaped reward function combined with joint limit clamping shows a significant improvement of $24\%$ over the baseline for small tolerance distances. Moreover, the proposed agent is able to solve whole-body control tasks which the baseline fails. We show that training the agent with our reward in one environment, transfers its learned skills well to a similar, but different, testing environment.

The current method is limited to using 2D LIDAR data and planar manipulator actions. Future work will focus on expanding these limitations via more informative observations, such as 3D LIDAR or RGB-D images.  More importantly, the ``folding arm'' behaviour should be further examined.

\section*{Acknowledgement}
We would like to thank Julien Kindle for his support in running the baseline agent simulation and training code.

\bibliographystyle{IEEEtran}
\bibliography{icra_2021_denis}

\section* {Appendix: Hyperparameters}

\begin{table}[!h]
    \centering
    \begin{tabular}{|c|c||c|c||c|c|}
        \hline
        \textbf{Parameter} & \textbf{Value} 
        & \textbf{Parameter} & \textbf{Value}
        & \textbf{Parameter} & \textbf{Value} \\\hline
        $w_{t}$ &  -15 & $d_{h}$  & 0.3 m &
        clip range & 0.2 \\\hline
        $w_{hd}$ &  40 & $D_h$ & 10  &
        clip range vf & -1\\\hline
        $w_{ht}$ & 20 & ${D_{jl}}^*$ & -20  &
        noptepochs & 30\\\hline
        $w_{pd}$ &  -10 & $D_c$ & -60  &
        gamma & 0.999\\\hline
        $w_{pt}$ &  50 & $\tau$ & 0.04 s  &
        n steps & 2048\\\hline
        ${w_{sm}}^*$ & -1 & $T_{h}$  & 1 s  &
        nminibatches & 8\\\hline
    \end{tabular}
    \caption{Reward parameters used for the reward and PPO agent. Parameters marked with $*$ are only used for the baseline}
    \label{tab:appendix:baseline-params}
\end{table}

\end{document}